\documentclass[letterpaper]{article} 
\usepackage{aaai23}  
\usepackage{times}  
\usepackage{helvet}  
\usepackage{courier}  
\usepackage[hyphens]{url}  
\usepackage{graphicx} 
\usepackage{natbib}  
\usepackage{caption} 
\usepackage{algorithm}
\usepackage[noend]{algpseudocode}

\title{Certifying Fairness of Probabilistic Circuits}
\author {
    Nikil Roashan Selvam\textsuperscript{\rm 1},
    Guy Van den Broeck\textsuperscript{\rm 1},
    YooJung Choi\textsuperscript{\rm 2 \thanks{This work was performed while YC was at UCLA.}}
}
\affiliations {
    \textsuperscript{\rm 1} Computer Science Department, University of California, Los Angeles\\
    \textsuperscript{\rm 2} School of Computing and Augmented Intelligence, Arizona State University\\
    nikilrselvam@cs.ucla.edu, guyvdb@cs.ucla.edu, yj.choi@asu.edu
}

\usepackage[utf8]{inputenc} 
\usepackage[T1]{fontenc}    
\usepackage{url}            
\usepackage{booktabs}       
\usepackage{amsfonts}       
\usepackage{nicefrac}       
\usepackage{microtype}      
\usepackage[usenames, dvipsnames]{xcolor}         
\usepackage{amsthm}         
\usepackage{xcolor}
\usepackage{xspace}
\usepackage{graphicx}

\usepackage{amsmath}
\usepackage{subcaption}
\usepackage[capitalise,nameinlink,noabbrev]{cleveref}
\usepackage{multirow}
\usepackage{wrapfig}

\usepackage{tikz}
\usetikzlibrary{spn}
\newcommand{\midlinewidth}{1.0pt}
\newcommand{\incmidlinewidth}{1.7pt}
\newcommand{\middist}{20pt}
\newcommand{\halfdist}{13pt}
\definecolor{petroil2} {RGB} {36, 165, 175}
\definecolor{gold2} {RGB} {255, 130, 0}

\newcommand{\PO}[1]{\ensuremath{#1_\text{PO}}}
\newcommand{\MN}[1]{\ensuremath{#1_\text{min}}}
\newcommand{\MX}[1]{\ensuremath{#1_\text{max}}}

\newcommand{\UB}{\mathit{UB}}

\DeclareMathOperator{\KL}{\ensuremath{D_{KL}}}

\newcommand{\PC}{\ensuremath{\mathcal{C}}}
\newcommand{\ch}{\ensuremath{\mathsf{ch}}}
\newcommand{\val}{\ensuremath{\mathsf{val}}}
\newcommand{\supp}{\ensuremath{\mathsf{supp}}}

\DeclareMathOperator*{\argmax}{arg\,max}
\DeclareMathOperator*{\argmin}{arg\,min}

\newcommand{\rvars}[1]{\ensuremath{\mathbf{#1}}\xspace}
\newcommand{\X}{\rvars{X}}
\newcommand{\Y}{\rvars{Y}}
\newcommand{\Z}{\rvars{Z}}
\newcommand{\E}{\rvars{E}}
\newcommand{\Ss}{\rvars{S}}
\newcommand{\Us}{\rvars{U}}
\newcommand{\Vs}{\rvars{V}}

\newcommand{\jstate}[1]{\ensuremath{\mathbf{#1}}\xspace}
\newcommand{\x}{\jstate{x}}
\newcommand{\y}{\jstate{y}}
\newcommand{\z}{\jstate{z}}
\newcommand{\us}{\jstate{u}}
\newcommand{\vs}{\jstate{v}}

\newcommand{\s}{\jstate{s}}

\newcommand{\abs}[1]{\left\lvert#1\right\rvert}

\theoremstyle{definition}

\newtheorem{defn}{Definition}
\newtheorem{lem}{Lemma}

\newcommand{\rethm}[3]{\newtheorem*{#1}{\cref{#1}}\begin{#1}[#2]#3\end{#1}}

\algnewcommand\algorithmicinput{\textbf{Input:}}
\algnewcommand\algorithmicoutput{\textbf{Output:}}
\algnewcommand\algorithmicdata{\textbf{Data:}}
\algnewcommand\Input{\item[\algorithmicinput]}%
\algnewcommand\Output{\item[\algorithmicoutput]}%
\algnewcommand\Data{\item[\algorithmicdata]}%
\algnewcommand{\IfThen}[2]{
\State \algorithmicif\ #1\ \algorithmicthen\ #2}

\newcommand{\eat}[1]{}

\begin{document}

\maketitle

\begin{abstract}
With the increased use of machine learning systems for decision making, questions about the fairness properties of such systems start to take center stage. Most existing work on algorithmic fairness assume complete observation of features at prediction time, as is the case for popular notions like statistical parity and equal opportunity. However, this is not sufficient for models that can make predictions with partial observation as we could miss patterns of bias and incorrectly certify a model to be fair.
To address this, a recently introduced notion of fairness asks whether the model exhibits any \textit{discrimination pattern}, in which an individual---characterized by (partial) feature observations---receives vastly different decisions merely by disclosing one or more sensitive attributes such as gender and race. By explicitly accounting for partial observations, this provides a much more fine-grained notion of fairness.
%
In this paper, we propose an algorithm to search for discrimination patterns in a general class of probabilistic models, namely probabilistic circuits. Previously, such algorithms were limited to naive Bayes classifiers which make strong independence assumptions; by contrast, probabilistic circuits provide a unifying framework for a wide range of tractable probabilistic models and can even be compiled from certain classes of Bayesian networks and probabilistic programs, making our method much more broadly applicable.
Furthermore, for an unfair model, it may be useful to quickly find discrimination patterns and distill them for better interpretability. As such, we also propose a sampling-based approach to more efficiently mine discrimination patterns, and introduce new classes of patterns such as minimal, maximal, and Pareto optimal patterns that can effectively summarize exponentially many discrimination patterns. 
\end{abstract}

\section{Introduction}

Machine learning systems are increasingly being used for critical decision making in a variety of areas ranging from education and health care, to financial lending and recidivism prediction~\citep{chouldechova2017fair,berk2018fairness,datta2015automated,henderson2015credit}. Consequently, there has been growing interest and concern about the fairness properties of these methods. In particular, biases in the training data and model architecture can result in certain individuals or groups receiving unfavorable treatment based on some sensitive attributes such as gender and race.
Naturally, various notions of fairness and ways to enforce them have been proposed ~\citep{barocas2016big,dwork2012fairness,hardt2016equality,nabi2018fair,madras2018fairness,salimi2019interventional}.

In this paper, we investigate the fairness properties of probabilistic models that represent joint distributions over the decision/prediction variable as well as the features. Such models are ubiquitous in decision-making systems for various real-world applications~\citep{koller2009probabilistic,sonnenberg1993markov,GRIFFITHS2010357}. In particular, they can be used to make classifications by inferring the probability of the class given some observations. Thus, by handling classifications as inference tasks, they can naturally handle missing features at prediction time.

While many existing work on algorithmic fairness assume that predictions are always made with complete observations of features, this fails to analyze the behavior of a model---in particular, its fairness---when making decisions with missing features. 
On the other hand, the notion of discrimination pattern~\citep{choi2020learning} explicitly aims to address fairness of decisions made with partial information. Specifically, it refers to an individual or a group of individuals, characterized by some partial assignments to the features, who may see a significant discrepancy in the prediction after additionally disclosing some sensitive attributes. 
While the complete observation case is also covered, it is merely a special case of discrimination patterns which in fact take into account all possible partial observation of features. As we show later, a model that is deemed fair according to popular notions such as disparate impact can still exhibit hundreds of discrimination patterns, when considering missing features.

This fine-grained notion leads to new challenges in certifying fairness, as there are now exponentially many patterns of unfairness to check and analyze. 
Thus, our first key contribution is to introduce special classes of discrimination patterns---minimal, maximal, and Pareto optimal patterns---which can ``summarize'' a large number of patterns. The set of these summary patterns are often far smaller than the set of all discrimination patterns, making them great targets to find in a model in order to discover and understand its unfairness.


The next contributions we make are algorithms to find discrimination patterns. 
The existing algorithm is limited to naive Bayes models, which make strong assumptions that may not suit real-world data.
On the other hand, our proposed methods can be applied to a more general class of models, namely probabilistic circuits (PCs)~\citep{ProbCirc20}, which have demonstrated competitive performance in various density estimation tasks~\citep{DangIJAR22,LiuICLR22,PeharzICML20}.
These are a family of probabilistic models that support tractable inference, encompassing arithmetic circuits~\citep{darwicheJACM-POLY}, sum-product networks~\citep{poon2011sum}, and-or graphs~\citep{DBLP:journals/ai/DechterM07}, probabilistic sentential decision diagrams~\citep{KisaVCD14}, and more. Some classical graphical models---such as those with bounded treewidth~\citep{ChowLiu} and their mixtures~\citep{meila2000learning}---as well as certain kinds of probabilistic programs~\citep{HoltzenOOPSLA20} can even be compiled into PCs, and thus can benefit from our proposed methods.

We propose a search-based algorithm that can find all discrimination patterns (or a special subset of them) in a PC or otherwise certify that there exists none and thus the PC is fair. Moreover, we introduce a sampling-based method, which can no longer prove the non-existence of discrimination patterns, but can far more efficiently find many of them in a PC, provided that they exist.
Through empirical evaluation on three benchmark datasets, we demonstrate that our search algorithm is able to find all discrimination patterns while only traversing a fraction of the space of possible patterns. Furthermore, we show that the sampling-based approach is indeed significantly faster in pattern mining, while still returning similar summary patterns as the exact approach.

\section{Discrimination Patterns}
\label{sec:disc-patterns}

\paragraph{Notation} We denote random variables by uppercase letters ($X$) and their assignments by lowercase letters ($x$). We use bold uppercase ($\X$) and lowercase letters ($\x$) for sets of variables and their assignments, respectively. The set of possible values of $\X$ is denoted by $\val(\X)$.
$D$ denotes a binary decision variable, and we use $d$ to refer to the assignment to $D$ that represents a favorable decision (e.g.\ a loan approval). We assume that a set of discrete (categorical) variables $\Z$ are used to make decisions. Furthermore, a subset of variables $\Ss \subset \Z$ are designated as \textit{sensitive attributes}, which are protected characteristics such as race or gender.

The notion of a discrimination pattern was introduced to study fairness of decisions made given partial observations of features.
They are particularly relevant for probabilistic models which can naturally handle missing value predictions, treating each prediction as an inference problem to compute the probability of a decision given some observations.

\begin{defn}[Discrimination patterns~\citep{choi2020learning}]\label{def:disc-pattern}
Let $P$ be a probability distribution over $D \cup \Z$, and $\x$ and $\y$ be joint assignments to $\X\subseteq\Ss$ and $\Y\subseteq\Z\setminus\X$, respectively. For some threshold $\delta \in [0,1]$, we say $\x$ and $\y$ form a \emph{discrimination pattern} w.r.t.\ $P$ and $\delta$ if:
\begin{equation*}
    \abs{ P(d \mid \x,\y) - P(d \mid \y) } > \delta.
\end{equation*}
We refer to the LHS as the \emph{discrimination score} of pattern $\x,\y$, denoted by $\Delta(\x,\y)$.
\end{defn}

Intuitively, a discrimination pattern corresponds to an individual (or a group of individuals sharing the same attributes) who would see a significant difference in the probability of getting a favorable decision, just by disclosing some sensitive information. Clearly, we want to avoid such scenarios; thus, the model is said to be $\delta$-fair iff it exhibits no discrimination patterns with respect to $\delta$.

\subsection{Relation to Other Fairness Notions}
\label{sec:related-fairness}

Before we describe our main contributions, let us briefly discuss how discrimination patterns relate to some other notions of fairness in the literature.
Many prominent fairness definitions---such as statistical/demographic parity (SP), disparate impact (DI), and equalized odds (EO)---fall under the notion of group fairness, which aims to equalize certain quantities across demographic groups~\citep{Feldman15,hardt2016equality}. While these definitions assume that predictions are made given complete observation of features, discrimination patterns account for fairness of decisions made with partial information, providing a more fine-grained notion of fairness. 
For instance, 
by definition a discrimination pattern of the form $(\s,\emptyset)$ means that $\abs{P(d\mid\s)-P(d)}>\delta$, which implies a violation of statistical parity.
That is, a $\delta$-fair model satisfies statistical parity for some threshold that depends on $\delta$, while the converse does not hold.
On the other hand, discrimination patterns formed by complete assignments can be interpreted as a violation of individual fairness~\citep{dwork2012fairness}; we refer to \citet{choi2020learning} for detailed examples.

Various methods have been proposed to verify different notions of fairness~\citep{Galhotra17,Bellamy18}, including those that utilize probability distributions~\citep{Albarghouthi17,Bastani19,Ghosh_Basu_Meel_2021,Ghosh_Basu_Meel_2022}. However, they often make simplifying assumptions such as conditional independence between attributes, making it challenging to scale these methods to probabilistic circuits which are more general as we will show later.
More importantly, verifying properties such as DI and SP is not sufficient if we wish to certify fairness according to discrimination patterns.

To illustrate this on real-world data, we learn PCs on the COMPAS dataset before and after fair data repair by \citet{Feldman15}. We use different degrees of data repair (controlled by parameter $\lambda$), yielding PCs with varying levels of fairness according to the above-mentioned standard notions (\cref{tab:fairness-comparison}). While the learned PCs would be certified as fair according to verifiers using metrics such as DI, SP, and EO, they may still exhibit discrimination patterns. For instance, many learned models in \cref{tab:fairness-comparison} satisfy SP with $\epsilon<0.02$ and EO with $\epsilon=0$, but still exhibit hundreds of discrimination patterns with scores $\delta>0.05$, some as high as $0.22$.

\begin{table}[t]
\caption{
Different metrics of fairness\protect\footnotemark for PCs on the COMPAS dataset before and after fair data repair~\citep{Feldman15}.
}

\label{tab:fairness-comparison}
\centering
\scalebox{0.75}{
\begin{tabular}{l|rrrrrr}
\toprule
                                  & Original & $\lambda\!=\!.5$  & $\lambda\!=\!.9$  & $\lambda\!=\!.95$ & $\lambda\!=\!.99$ & $\lambda\!=\!1   $ \\\midrule
DI                                    & 0.187    & 0.063  & 0.017  & 0.020  & 0.015  & 0.023  \\ 
SP                                    & 0.183    & 0.061  & 0.016  & 0.019  & 0.014  & 0.022  \\
SP (1 variable)                       & 0.055   & 0.015 & 0.002 & 0.001 & 0.001 & 0.001 \\ 
EO                                    & 0.752    & 0.000    & 0.000    & 0.000    & 0.000    & 0.000    \\ 
\# Disc.\ Patt.\ ($0.05$) & 3866        & 1320      & 488      & 659      & 894      & 578      \\ 
\# Disc.\ Patt.\ ($0.1$)  & 1761     & 311    & 11     & 64     & 74     & 34     \\ 
Highest Disc.\ Score     & 0.372    & 0.208  & 0.112  & 0.225  & 0.176  & 0.123  \\
\bottomrule
\end{tabular}}
\end{table}

\footnotetext{DI $= 1-\frac{\min _{\s} \mathbb{P}(d \mid \s)}{\max_{\s} \mathbb{P}(d \mid \s)}$,\; SP$= \max _{\s} \mathbb{P}(d \mid \s)-\min_{\s} \mathbb{P}(d \mid \s)$,\\ SP (1 variable) $= \max _{s} \mathbb{P}(d \mid s)-\min _{s} \mathbb{P}(d \mid s)$ for a single $s$, \\ EO $= \max \{ \max _{\s} \mathbb{P}(\hat{d} \mid \s d)-\min_{\s} \mathbb{P}(\hat{d} \mid \s d), \max _{\s} \mathbb{P}(\hat{d} \mid \s\overline{d})-\min_{\s} \mathbb{P}(\hat{d} \mid \s\overline{d}) \}$ .}



That is, discrimination patterns enable a more fine-grained auditing of fairness. However, even a simple classifier could exhibit a large number of discrimination patterns, thereby making it hard for domain experts and users to examine them effectively. For instance, a naive Bayes classifier with 7 features for the COMPAS data\footnote{\url{https://github.com/propublica/compas-analysis}} was shown to have more than 2,000 discrimination patterns~\citep{choi2020learning}; this is clearly not scalable for interpretation.
Thus, we propose new classes of discrimination patterns that can be used as representatives for a large number of patterns, thereby being more amenable for interpretations.

\subsection{Summarizing Patterns}
\label{sec:summarizing}


A natural way to choose the most ``interesting'' patterns may be by ranking them by their discrimination scores, and focusing on a few instances that are the most discriminatory. While it may be useful to study the most problematic patterns and address them, they do not necessarily provide insight into other discrimination patterns that exist.
Instead, we propose the notion of maximal and minimal patterns that can summarize groups of patterns, namely their extensions and contractions.
An extension of a pattern $(\x,\y)$, denoted by $(\x',\y') \supset (\x,\y)$, can be generated by adding an assignment to the pattern: that is, $\x\subseteq\x'$, $\y\subseteq\y'$, and either $\x\!\subset\!\x'$ or $\y\!\subset\!\y'$. Conversely, $(\x,\y)$ is called a contraction of $(\x',\y')$.

\begin{defn}[Maximal patterns]
Let $\Sigma$ denote a set of discrimination patterns w.r.t.\ a distribution $P$ and threshold $\delta$. The set of \emph{maximal patterns} $\MX{\Sigma}\subseteq\Sigma$ consists of all patterns $(\x,\y)\in\Sigma$ that are not a complete assignment (i.e.\ $\Z \setminus (\X\cup\Y) \not=\emptyset$)
and:
\begin{equation*}
    \forall (\x',\y')\supset(\x,\y),\ (\x',\y')\not\in\Sigma.
\end{equation*}
\end{defn}

In other words, a maximal pattern is a discrimination pattern such that none of its extensions are discrimination patterns. As the name suggests, an extension of a maximal pattern cannot also be maximal, because by definition it will not be a discrimination pattern.
Hence, an individual with attributes $\x$ and $\y$ who may see a discrimination in the decision by disclosing their sensitive information would no longer receive such treatment if they additionally share other features, whatever their values may be.
This notion is nicely complemented by the following notion of minimal patterns.


\begin{defn}[Minimal patterns]
Let $\Sigma$ denote a set of discrimination patterns w.r.t.\ a distribution $P$ and threshold $\delta$. The set of \emph{minimal patterns} $\MN{\Sigma}\subseteq\Sigma$ consists of all patterns $(\x,\y)\in\Sigma$ such that:
\begin{align*}
    & \forall\, (\x',\y') \supset (\x,\y),\  (\x',\y') \in \Sigma \\ \quad\text{and}\quad
    & \forall\, (\x'',\y'') \subset (\x,\y)\text{ s.t. }\x''\not=\emptyset,\  (\x'',\y'') \not\in \Sigma.
\end{align*}
\end{defn}

That is, we say a discrimination pattern is minimal if all of its extensions and none of its contractions are discriminatory. Thus, a single minimal pattern can be interpreted as representing a large set of discrimination patterns---more precisely, whose size is exponential in the number of unobserved features. 
For example, suppose $\Z=\{X,Y,U\}$ are binary features with $\Ss=\{X\}$, and let $(\{X\!=\!1\},\{Y\!=\!0\})$ be a minimal pattern. Then its valid contraction---i.e.\ $(\{X\!=\!1\},\{\})$---is not a discrimination pattern; while all of its extensions---i.e.\ $(\{X\!=\!1\},\{Y\!=\!0,U\!=\!0\})$ and $(\{X\!=\!1\},\{Y\!=\!0,U\!=\!1\})$---are discriminatory. Note that by definition a minimal pattern cannot be a maximal pattern, and vice versa.


As a case study, we train a probabilistic model\footnote{See \cref{sec:exp} for details of the trained model.} on the COMPAS dataset, which exhibits 7445, 2338, and 1164 discrimination patterns for $\delta=0.01$, $0.05$, and $0.1$ respectively. On the other hand, this model has 170, 74, and 0 maximal patterns, and only 103, 10, and 1 minimal patterns, for respective values of threshold $\delta$.
%
Interestingly, we observe that among the 74 maximal patterns for $\delta=0.05$, none of them includes an assignment to the variable regarding `supervision level', suggesting that there are many instances where an individual would not see an unfair prediction if the supervision level is additionally known.
Moreover, remarkably, the single minimal pattern in the case of $\delta=0.1$ can represent 512 patterns (its extensions) out of 1164 total discrimination patterns in the model.\footnote{The minimal pattern is where $\x=\{\text{Not Married}\}$ and $\y=\{\text{Low--Medium Supervision Level}\}$.}

Another important consideration when studying discrimination patterns is their probability. Recall that each pattern $(\x,\y)$ represents a group of people sharing the attributes $\x$ and $\y$. Then the probability $P(\x,\y)$ corresponds to the proportion of the population that could be affected. 
Even though any unfairness is equally undesirable for minority groups (i.e. lower probability) as it is for majority groups, patterns that are so specific and have exceedingly low probability would not be as insightful or even relevant when auditing a model for real-world fairness concerns.

To address this, patterns can be ranked by their divergence score, which takes into account the probability of a pattern as well as its discrimination score.

\begin{defn}[Divergence score~\citep{choi2020learning}]
Let $P$ be a probability distribution over $D \cup \Z$ and $\delta$ some threshold in $[0.1]$. Further suppose $\x$ and $\y$ are joint assignments to $\X\subseteq\Ss$ and $\Y\subseteq\Z\setminus\X$, respectively. 
Then the \emph{divergence score} of $(\x,\y)$ is:
\begin{align*}
    \min_Q\: \KL\left(P \;\middle\|\; Q\right)
    \text{ s.t. } & \Delta(\x,\y) \leq \delta,\; \\
     & P(d,\z) = Q(d,\z), \:\forall\: \z \not\supset \x\cup\y 
\end{align*}
where $\KL\left(P\;\middle\|\; Q\right) = \sum_{d,\z} P(d,\z) \log(P(d,\z)/Q(d,\z))$.
\end{defn}

Informally, it aims to quantify how much the distribution $P$ needs to be changed in order to remove the discrimination pattern $(\x,\y)$. Thus, the patterns with highest divergence scores would tend to have both high discrimination score as well as high probability.
In fact, we could summarize all existing discrimination patterns in a model by explicitly constructing a set of patterns with the following behavior: one cannot increase the discrimination score without decreasing the probability, and vice versa.

\begin{defn}[Pareto optimal patterns]
Let $\Sigma$ denote a set of discrimination patterns w.r.t.\ a distribution $P$ and threshold $\delta$. The set of \emph{Pareto optimal patterns} $\PO{\Sigma} \subseteq \Sigma$ consists of the patterns $(\x,\y)\in\Sigma$ such that:
\begin{align*}
    &\forall (\x',\y')\in\Sigma\setminus\{(\x,\y)\},\quad \\
    & \Delta(\x,\y) > \Delta(\x',\y') \,\text{ or }\, P(\x,\y) > P(\x',\y').
\end{align*}
\end{defn}

Pareto optimal patterns can be a very effective way to study fairness of a probabilistic model, as it significantly reduces the number of discrimination patterns one would examine. For example, the model trained on the COMPAS with 2388 and 1164 discrimination patterns w.r.t.\ $\delta=0.05$ and $0.1$, respectively, has only 38 and 28 Pareto optimal patterns.
That is, for $\delta=0.1$, each of the $1164-28=1136$ patterns has discrimination score and probability that are both dominated by those of some Pareto optimal pattern.

In the following sections, we discuss how to find discrimination patterns in probabilistic circuits, as well as generating the summaries in the form of maximal, minimal, and Pareto optimal patterns.

\section{Finding Discrimination Patterns in Probabilistic Circuits}
\label{sec:search}

We now describe our algorithm to find discrimination patterns, if there exists any, or certify that there are none.
As discussed in \cref{sec:summarizing}, the probability of a pattern corresponds to the proportion of the affected subpopulation, according to the probabilistic model. Therefore, a meaningful analysis of discrimination patterns depends on how well the model captures the population distribution.
For instance, the existing algorithm to discover discrimination patterns assumes naive Bayes classifiers,
which make strong independence assumptions and are generally too restrictive to fit real-world distributions.
We instead consider a more expressive type of probabilistic models, called probabilistic~circuits.

\subsection{Probabilistic Circuits}


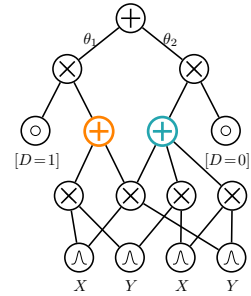
\begin{figure}[tb]
\centering
\scalebox{0.6}{
\begin{tikzpicture}
	\sumnode[line width=\midlinewidth]{s1};
	\prodnode[line width=\midlinewidth, below=\halfdist of s1, xshift=-2*\middist]{p1};
	\prodnode[line width=\midlinewidth, below=\halfdist of s1, xshift=2*\middist]{p2};

    \bernode[line width=\midlinewidth, below=\middist of p1, xshift=-\middist]{v1}{$[D\!=\!1]$};
	\bernode[line width=\midlinewidth, below=\middist of p2, xshift=\middist]{v2}{$[D\!=\!0]$};

    \sumnode[line width=\incmidlinewidth, below=\middist of p1, xshift=\middist, draw=gold2]{s2};
    \sumnode[line width=\incmidlinewidth, below=\middist of p2, xshift=-\middist, draw=petroil2]{s3};
    
    \prodnode[line width=\midlinewidth, below=4*\middist+\halfdist of s1]{p3};
    \prodnode[line width=\midlinewidth, left=\middist of p3]{p4};
    \prodnode[line width=\midlinewidth, right=\halfdist of p3]{p5};
    \prodnode[line width=\midlinewidth, right=\halfdist of p5]{p6};

	\contnode[line width=\midlinewidth, below=\middist of p4, xshift=0.5*\halfdist]{v3}{$X$};
	\contnode[line width=\midlinewidth, right=\halfdist of v3]{v4}{$Y$};
	\contnode[line width=\midlinewidth, right=\halfdist of v4]{v5}{$X$};
	\contnode[line width=\midlinewidth, right=\halfdist of v5]{v6}{$Y$};

  \weigedge[line width=\midlinewidth,left,pos=0.37] {s1} {p1} {$\theta_1$};
  \weigedge[line width=\midlinewidth,right,pos=0.37] {s1} {p2} {$\theta_2$};

  \edge[line width=\midlinewidth,right,>=stealth] {p1} {v1};
  \edge[line width=\midlinewidth,left,>=stealth] {p2} {v2};
  \edge[line width=\midlinewidth,left,>=stealth] {p1} {s2};
  \edge[line width=\midlinewidth,left,>=stealth] {p2} {s3};
  
  \edge[line width=\midlinewidth,left,>=stealth] {s2} {p3};
  \edge[line width=\midlinewidth,left,>=stealth] {s2} {p4};
  \edge[line width=\midlinewidth,left,>=stealth] {s3} {p3};
  \edge[line width=\midlinewidth,left,>=stealth] {s3} {p5};
  \edge[line width=\midlinewidth,left,>=stealth] {s3} {p6};
  
  \edge[line width=\midlinewidth,left,>=stealth] {p4} {v3};
  \edge[line width=\midlinewidth,left,>=stealth] {p4} {v4};
  \edge[line width=\midlinewidth,left,>=stealth] {p3} {v3};
  \edge[line width=\midlinewidth,left,>=stealth] {p3} {v6};
  \edge[line width=\midlinewidth,left,>=stealth] {p5} {v4};
  \edge[line width=\midlinewidth,left,>=stealth] {p5} {v5};
  \edge[line width=\midlinewidth,left,>=stealth] {p6} {v5};
  \edge[line width=\midlinewidth,left,>=stealth] {p6} {v6};
\end{tikzpicture}
}
\caption{A PC over variables $\{D,X,Y\}$}
\label{fig:pc}    
\end{figure}  

A probabilistic circuit is a directed acyclic graph (DAG) with parameters, in which inner nodes can either be sum or product nodes, and input nodes are associated with simple univariate distributions (often indicator functions in the case of discrete variables). Moreover, each edge $(n,c)$ between a sum node $n$ and its child $c$ is associated with a parameter $\theta_{n,c} > 0$. A PC $\PC$ over random variables $\X$ recursively defines a probability distribution over $\X$ as follows:
\begin{equation*}
n(\x)=
\begin{cases}
f_n(\x) &\text{if $n$ is a leaf node} \\
\prod_{c\in\ch(n)} c(\x) &\text{if $n$ is a product} \\
\sum_{c\in\ch(n)} \theta_{n,c} \cdot c(\x) &\text{if $n$ is a sum}
\end{cases}
\end{equation*}
Here, $\ch(n)$ denotes the set of children of an inner node $n$. Then the distribution $P$ defined by a PC is exactly $P(\x) = n(\x)$ where $n$ is the root of the PC.

A key strength of PCs is that they allow tractable inference of certain probabilistic queries, based on which structural properties are satisfied by the circuit. The first inference task we need is computing conditional probabilities, which is required to get the discrimination score (\cref{def:disc-pattern}) of a pattern. Moreover, we would also like to compute probabilities of patterns---that is, marginal probabilities given some partial observations. PCs support efficient marginal and conditional inference if they satisfy two structural properties called smoothness and decomposability.
A PC is \emph{smooth} if for every sum node its children include exactly the same set of variables, and it is \emph{decomposable} if for every product node its children depend on disjoint sets of variables~\citep{darwiche2002knowledge}. Given these properties, computing any marginal probability can be done through a single feedforward evaluation of the PC, thus taking linear time in the size of the circuit. Hence, we will assume smooth and decomposable PCs throughout this paper.

Let us quickly remark on the wide applicability of probabilistic circuits and subsequently our pattern mining algorithm which takes PCs and inputs. First, both the structure and parameters of PCs can be learned to fit the data, and in a wide range of 
tasks, they were shown to achieve competitive and state-of-the-art performance~\citep{DangIJAR22,LiuICLR22,PeharzICML20,LiUAI21}.
In addition, as mentioned previously, they can be compiled efficiently from bounded-treewidth graphical models, and thus we can apply the following search algorithm on such models as well.

\subsection{Search Algorithm}
\label{sec:search-algorithm}

To certify whether a distribution defined by a probabilistic circuit is $\delta$-fair, we search for discrimination patterns in the PC. If the search concludes without finding any pattern, then we know that the PC is $\delta$-fair; otherwise, we return all or some of the discrimination patterns selected according to one of the criteria discussed in \cref{sec:disc-patterns}.

\begin{algorithm}[!t]
\caption{$\textsc{Search-Disc-Patt}(\x,\y,\E)$}
\label{alg:search}
\begin{algorithmic}[1]
\Input{a PC $\PC$ over variables $D\cup\Z$ and a threshold $\delta$}
\Output{a set of discrimination patterns $\Sigma$}
\Data{current pattern $(\x,\y)\leftarrow(\{\},\{\})$; excluded variables $\E\leftarrow\{\}$}
\State{$\Sigma \leftarrow \{\}$}
\For{\textbf{each} $z\in\val(Z)$ for some $Z\in\Z\setminus (\X\cup\Y\cup\E)$}
\If{$Z\in\Ss$} 
    \If{$\Delta(\x\cup\{z\},\y) > \delta$}
        \State{$\Sigma \leftarrow \Sigma \cup \{(\x\cup\{z\},\y)\}$}
    \EndIf
    \If{$\UB(\x\cup\{z\},\y,\E) >  \delta$} 
        \State{$\Sigma\! \leftarrow \!\Sigma \cup \textsc{Search-Disc-Patt}(\x\!\cup\!\{\z\},\y,\E)$}
    \EndIf
\EndIf
\IfThen{$\Delta(\x,\y\cup\{z\}) > \delta$}{$\Sigma \leftarrow \Sigma \cup \{(\x,\y\cup\{z\})\}$} 
\If{$\UB(\x,\y\cup\{z\},\E) > \delta$} 
    \State{$\Sigma \leftarrow \Sigma \cup \textsc{Search-Disc-Patt}(\x,\y\cup\{z\},\E)$}
\EndIf
\EndFor
\If{$\UB(\x,\y,\E\cup\{Z\}) > \delta$} 
    \State{$\Sigma \leftarrow \Sigma \cup \textsc{Search-Disc-Patt}(\x,\y,\E\cup\{Z\})$}
\EndIf
\State{\Return $\Sigma$}
\end{algorithmic}
\end{algorithm}

More precisely, we adopt a branch-and-bound search approach. \cref{alg:search} outlines the pseudocode of our search algorithm. At each search step, it checks whether the current assignments form a discrimination pattern, and explores extensions by recursively adding variable assignments.
Note that while we mainly present the algorithm that returns all discrimination patterns, we can easily tweak it to return the top-k most discriminating patterns: by keeping a running list of top-k patterns and using the k-th highest score as the threshold instead of $\delta$.

As there are exponentially many potential patterns, we rely on a good upper bound to effectively prune the search tree. In particular, we use the following as our bound $\UB(\x,\y,\E)$:
\begin{align}\label{eq:ub}
\max \{ & \abs{\max_{\us} P(d\mid\x,\y,\us) - \min_{\us} P(d\mid\y,\us) }, \\ &\abs{\min_{\us} P(d\mid\x,\y,\us) - \max_{\us} P(d\mid\y,\us)} \}
\end{align}
where $\Us$ can be any subset of $\Z \setminus (\X\cup\Y\cup\E)$---in other words, the remaining variables to extend the current pattern.
The core component of above bound is maximizing or minimizing the conditional probability of the form $P(d\mid \y,\us)$ over the values of some $\Us$ for a given $\y$. We now show how such optimization can be done tractably for certain classes of probabilistic circuits.

We use two key observations, expressed by the following lemmas.\footnote{Complete proofs of lemmas can be found in the appendix.}
\begin{lem}\label{lem:complete}
Let $P$ be a distribution over $D\cup\Z$ and $\x$ a joint assignment to $\X\subseteq\Z$. Also denote $\Vs = \Z\setminus\X$. Then for any $\Us \subseteq \Z\setminus\X$ the following holds:
\begin{equation*}
    \max_{\us\in\val(\Us)} P(d \mid \x,\us) \leq \max_{\vs\in\val(\Vs)} P(d \mid \x,\vs)
\end{equation*}
\end{lem}
That is, to maximize a conditional probability given some (partial) assignments for a set of free variables, it suffices to consider only the complete assignments to those variables. Analogously, this statement holds for minimization as well, with the direction of inequality reversed.

\begin{lem}\label{lem:ratio}
Let $P$ be a distribution over $D\cup\Z$, $\x$ an assignment to $\X\subseteq\Z$, and $\Us\subseteq\Z\setminus\X$. Then,
\begin{equation*}
    \argmax_{\us\in\val(\Us)} P(d \mid \x,\us) = \argmax_{\us\in\val(\Us)} \frac{P(\x,\us \mid d)}{P(\x,\us \mid \overline{d})}.
\end{equation*}
\end{lem}
Combining these observations, we see that the upper bound in \cref{eq:ub} can be computed easily if we can efficiently maximize and minimize quantities of the form $P(\x,\us\mid d) / P(\x,\us\mid \overline{d})$.
In fact, we derive an algorithm with worst-case quadratic time complexity (in the size of the circuit) for PCs that satisfy additional structural constraints. Deferring the algorithmic details and proof of correctness to the appendix, here we instead provide high-level insights to these additional tractability conditions.
First, \citet{VergariNeurIPS21} shows the necessary structural conditions (called compatibility and determinism) such that the quotient of two PCs can be computed tractably and represented as another circuit representation that allows for linear-time optimization~\citep{ChoiDarwiche17}.
Then all we need is to represent the conditional distributions $P(\Z \mid d)$ and $P(\Z \mid \overline{d})$ as two PCs satisfying those conditions.
If the decision variable $D$ appears at the top of the PC over $D\cup\Z$ as illustrated in \cref{fig:pc}, then the two subcircuits rooted at each child of the root node exactly corresponds to the conditional distributions given $D$. For instance, the PCs rooted at the orange and green sum nodes in \cref{fig:pc} correspond to the conditional distributions $P(X,Y \mid D\!=\!1)$ and $P(X,Y \mid D\!=\!0)$, respectively.

Furthermore, we can similarly search for top-k patterns ranked by their divergence scores. \citet{choi2020learning} provides an upper bound on divergence score that once again requires efficiently maximizing/minimizing conditional probability of extensions. Thus, given the kinds of PC structure described above, we can also compute a non-trivial bound and extend the branch-and-bound search approach in a straightforward manner to mine divergence patterns in PCs as well. 

Lastly, we briefly discuss how to obtain the special types of discrimination patterns introduced in \cref{sec:summarizing}. 
We can make a small tweak to the search algorithm to keep track of the Pareto front of discrimination patterns found so far in each search step. 
Concretely, we maintain an ordered container storing the probability and discrimination score in increasing order of the former and decreasing order of the latter. 
Similarly, finding the set of maximal patterns is almost identical to searching for all discrimination patterns. In particular, when exploring a pattern in the search tree, we declare it to be maximal if no extension of it can be a discrimination pattern, determined by the quadratic-time upper bound on discrimination score.
For minimal patterns, we derive a sub-quadratic time algorithm to examine a set of patterns and recover the minimal ones; see appendix for details.



\section{Discrimination Pattern Mining by Sampling}
\label{sec:sampling}

\begin{algorithm}[!t]
\caption{$\textsc{Sample-Disc-Patterns}(\PC,\Z)$} \label{alg:sampling}
\begin{algorithmic}[1]
\Input{a PC $\PC$ over variables $D\cup\Z$ and a threshold $\delta$}
\Output{a set of sampled discrimination patterns $\Sigma$}
\State{$\Sigma \leftarrow \{\}$}
\Repeat \Comment{generate samples until timeout}
    \State{$(\x,\y)\leftarrow(\{\},\{\})$}
    \While{$\abs{\x}+\abs{\y} < n$}
        \For{$(\x',\y') \in \mathsf{extensions}(\x,\y))$}  \State{\Comment{each extension by a single variable}}
        \IfThen{$\Delta(\x',\y')>\delta$}{$\Sigma \leftarrow \Sigma \cup \{(\x',\y')\}$}
        \EndFor
        \State{$(\x,\y) \gets \mathsf{sample}_{\text{weight}:\Delta(\x'\y')}(\mathsf{extensions}(\x,\y))$} 
    \EndWhile
\Until{timeout}
\State{\Return $\Sigma$}
\end{algorithmic}
\end{algorithm}

Certifying that there exists no discrimination pattern, among exponentially many possible assignments, is a very hard problem. In real-world settings, one may simply be interested in quickly studying examples of unfairness that may be present in the model.
In fact, this is the goal for many existing fairness auditing tools~\citep{saleiro2018aequitas}: to find patterns of bias (potentially using different fairness definitions) for the developer or user to examine.
Hence, we introduce efficient sampling-based methods to mine discrimination patterns in PCs. While these methods cannot necessarily certify a model to be $\delta$-fair, they can very quickly find a large number of patterns, as we will later show empirically.


Our proposed approach is summarized in \cref{alg:sampling}. At a high level, each run of the sampling algorithm starts from an empty assignment $\x\y$ and incrementally adds one attribute at a time until a complete assignment is obtained. The attribute to be added (or more precisely the immediate extension to be explored) at each step is sampled at random with a likelihood proportional to the discrimination score of the resulting assignment. Any assignment explored along the way with a sufficiently high discrimination score is added to our set of patterns. Intuitively, at any assignment, we use the discrimination score of its immediate extension as a heuristic for its extensions being discrimination patterns. 

\cref{alg:sampling} is the base of our proposed sampling algorithm. We also derive a more sophisticated algorithm with memoization between samples and control over exploration versus exploitation at different stages of sampling. At a high level, we maintain an estimator at each assignment corresponding to the expected discrimination score of an extension and backtrack at the end of each sampling run to update the estimates before the next run. We refer the reader to the appendix for details regarding the same. Observe that the sampling algorithm is computationally inexpensive overall as the only circuit evaluations that need to be performed are a few feed-forward evaluations (linear time) to compute conditionals at each assignment explored. Lastly, it is worth noting that after the discrimination patterns have been sampled, we can utilize similar techniques as described in \cref{sec:search-algorithm} to efficiently summarize them through minimal, maximal, and Pareto optimal patterns.

\section{Empirical Evaluation}
\label{sec:exp}

We evaluate our discrimination pattern mining algorithms on three datasets: \textit{COMPAS} which is used for recidivism prediction and the \textit{Adult}~\citep{Dua:2019} and \textit{Income}~\citep{ding2021retiring} datasets for predicting income levels. As pre-processing, we remove redundant features and features with unique values, and discretize numerical values.
We learn a PC from each dataset using the \textsc{Strudel} algorithm~\citep{DangIJAR22}, which returns deterministic and structured decomposable PCs as required by our search algorithm.\footnote{Link to pre-processed data, trained models, and code: https://github.com/UCLA-StarAI/PC-DiscriminationPatterns.}
All experiments were run on an Intel(R) Xeon(R) CPU E5-2640 (2.40GHz).

With regards to empirical comparison with existing work in literature, it is worth emphasizing that efficiently mining discrimination patterns has only been possible so far for naive Bayes ~\citep{choi2020learning} which makes strong assumptions, and our method is (to the best of our knowledge) the first method to extend this to a much more general class of models, namely PCs. Furthermore, other existing fairness verifiers in literature as discussed in \cref{sec:related-fairness} are not able to find discrimination patterns: they only verify a weaker fairness properties such as statistical parity. Hence, in this section, we evaluate our exact and approximate methods against the only available baseline of naive enumeration, and against each other to clearly test comparative efficiency.


\subsection{Exact Search}

\begin{table}[t]
\caption{Dataset statistics (number of examples, number of sensitive features $S$, non-sensitive features $N$)
and speedup of top-k search v.s.\ naive enumeration, in terms of the fraction of search space explored.}
\label{tab:exact}
\centering
\scalebox{0.7}{
\begin{tabular}{llll|l|r|rrr}
\toprule
    &        &  &   &     &        \multicolumn{1}{c|}{Disc.}    &    \multicolumn{3}{c}{Divergence} \\
Dataset & Size   & $S$  & $N$ & $k$   & $\delta\!=\!0.1$ & $\delta\!=\!0.01$      & $\delta\!=\!0.05$           & $\delta\!=\!0.10$      \\ \midrule
\multirow{3}{*}{COMPAS}  & \multirow{3}{*}{48834} & \multirow{3}{*}{4}  & \multirow{3}{*}{3} & 1    & 2.73x  & 2.17x & 1.40x  & 1.16x \\
    &        &         &   &     10  & 2.68x & 1.85x & 1.26x  & 1.10x \\
    &        &         &   &     100 & 2.52x & 1.46x & 1.13x  & 1.04x \\ \midrule
\multirow{3}{*}{Income}  & \multirow{3}{*}{195665} & \multirow{3}{*}{2}       & \multirow{3}{*}{6} & 1   & 1.22x & 1.50x & 1.32x  & 1.13x \\
    &        &         &   &     10  & 1.20x & 1.40x & 1.26x  & 1.08x \\
    &        &         &   &     100 & 1.13x & 1.31x & 1.15x  & 1.02x \\ \midrule
\multirow{3}{*}{Adult}   & \multirow{3}{*}{32561}  & \multirow{3}{*}{4} & \multirow{3}{*}{9} &  1   & 1.32x & 24.20x & 16.72x  & 10.88x \\
    &        &         &   &     10 & 1.31x & 20.44x & 14.75x  & 9.82x \\
    &        &         &   &     100 & 1.29x & 16.10x & 11.87x  & 8.40x \\ \bottomrule 
\end{tabular}}
\end{table}

We first evaluate the efficiency of our branch-and-bound search algorithm to find discrimination patterns. As our approach is the first non-trivial method that does not require naive Bayes assumption,
we see whether it is more efficient than a naive solution that enumerates all possible patterns.
%
We mine the top-k patterns for two ranking heuristics (discrimination and divergence score), three values of $k$ (1, 10, 100), and three threshold values $\delta$ (0.01, 0.05, 0.1).
\cref{tab:exact} reports the speedup in terms of the proportion of the search space visited by our algorithm compared to the naive approach. 
Note that only the settings in which $\delta\!=\!0.1$ for ranking by discrimination score are reported, because the results are identical for smaller values of $\delta$.
We observe that pruning is effective, resulting in consistent speedup, including some significant improvement in performance as high as 24x speedup in the case of mining top-k divergence patterns on the Adult dataset.

Moreover, note that our method computes an upper bound at every search step, which has a worst-case quadratic time complexity. However, we see that pruning the search space still improves the overall run time of the algorithm, even with this extra computation. For example, our method explores a little less than half the search space for top-k discrimination patterns with $\delta\!=\!0.1$ on the COMPAS dataset, and it takes about 60\% of the time taken by naive enumeration; concretely, it takes 24.4s, 24.6s, and 25.3s for $k=1,10,100$, respectively, while the naive approach takes 40.2s.

\subsection{Sampling}

\begin{table}[t]
\caption{Average speedup of sampling v.s.\ naive enumeration to find top 1 pattern
(in terms of proportion of search space explored).}
\label{tab:sampling_expectation}
\centering
\scalebox{0.8}{
\begin{tabular}{l|rr|rr}
\toprule
& \multicolumn{2}{c|}{Discrimination} & \multicolumn{2}{c}{Divergence} \\
Dataset & Avg     & StdDev  & Avg & StdDev \\ \midrule
Compas  & 26x   & 54x    & 29x    & 15x    \\ 
Income  & 17x   & 14x  & 143x    & 6x     \\ 
Adult   & 48x & 53x & 49480x     & 2113x     \\  \bottomrule
\end{tabular}}
\end{table}

The primary motivation for the sampling algorithm is to not only quickly audit a model to check if it exhibits any discrimination patterns, but in particular to quickly analyze the most interesting patterns. To that end, we first evaluate how efficiently the sampling algorithm is able to find the most discriminatory/divergent pattern. 
We first find the highest discrimination and divergence score using exact search, then run the sampling algorithm until it finds the top-1 pattern (using $\delta=0.01$ for the divergence scores).
\cref{tab:sampling_expectation} details the speedup relative to naive enumeration, in terms of the number of assignments explored to find the top-1 pattern; each result is the average over 10 independent random trials. 


Recall that each sampling instance merely requires a few feed-forward evaluations of the circuit (linear time) for computing marginals. Thus, from the table it is clear that the sampling algorithm is much quicker than exact search in finding the patterns with highest scores (for instance, as much as 49480x fewer patterns explored compared to naive search to find the most divergent pattern in Adult). 

\begin{table}[t]
\caption{Number of discrimination patterns and highest score found by exact search and sampling.}
\label{tab:timeout}
\centering
\scalebox{0.8}{
\begin{tabular}{lrc|rr|rr}
\toprule
&           &       &   \multicolumn{2}{c|}{\# Patterns found}    & \multicolumn{2}{c}{Highest score} \\
Dataset & Time   & $\delta$ & Exact & Sampling & Exact & Sampling \\ \midrule
\multirow{2}{*}{COMPAS}  & \multirow{2}{*}{3s} & 0.05 & 347 & \textbf{751} & \textbf{0.2236} & 0.2230 \\
&    & 0.10 & 210 & \textbf{347} & \textbf{0.2236} & 0.2230 \\ \midrule
\multirow{2}{*}{Income} & \multirow{2}{*}{5s}  & 0.05 & 209 & \textbf{1090} & 0.1076 & \textbf{0.1658} \\
&    & 0.10 & 3 & \textbf{225} & 0.1076 & \textbf{0.1659} \\ \midrule
\multirow{2}{*}{Adult}  & \multirow{2}{*}{600s} & 0.05 & 37167 & \textbf{113763} & 0.6725 & \textbf{0.6871} \\
&    & 0.10 & 30982 & \textbf{99578} & 0.6725 & \textbf{0.6844} \\ \bottomrule
\end{tabular}}
\end{table}

For a more direct comparison, we run the exact search algorithm and the sampling method with a specified timeout. We report both the number of patterns found and the highest score in different settings (\cref{tab:timeout}). We observe that the sampling algorithm consistently outperforms the exact search in both number of patterns found the scores of patterns found.
Thus, we can reliably use the sampling approach to quickly mine many discrimination patterns of significance.


\begin{figure}[tb]
\centering
\begin{subfigure}{0.25\textwidth}
\centering
\includegraphics[width=0.94\columnwidth]{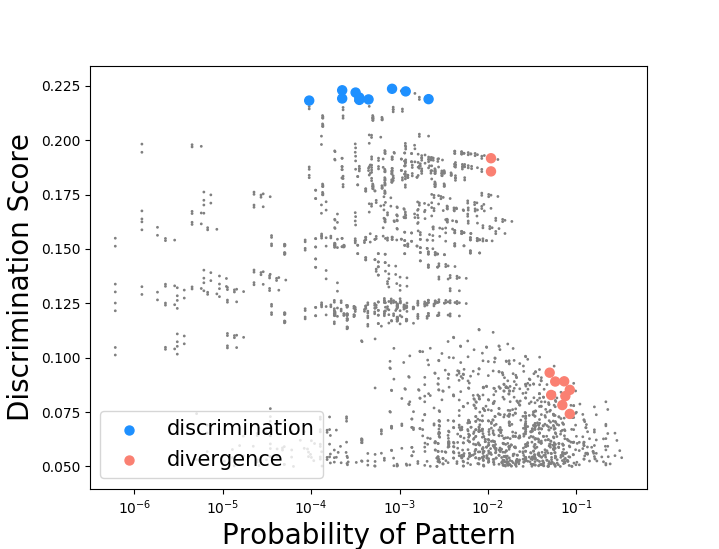}
\caption{Top 10 discrimination and divergence patterns found by sampling.}
\label{fig:top10sampling}
\end{subfigure}
\\
\begin{subfigure}{0.23\textwidth}
\centering
\includegraphics[width=0.98\columnwidth]{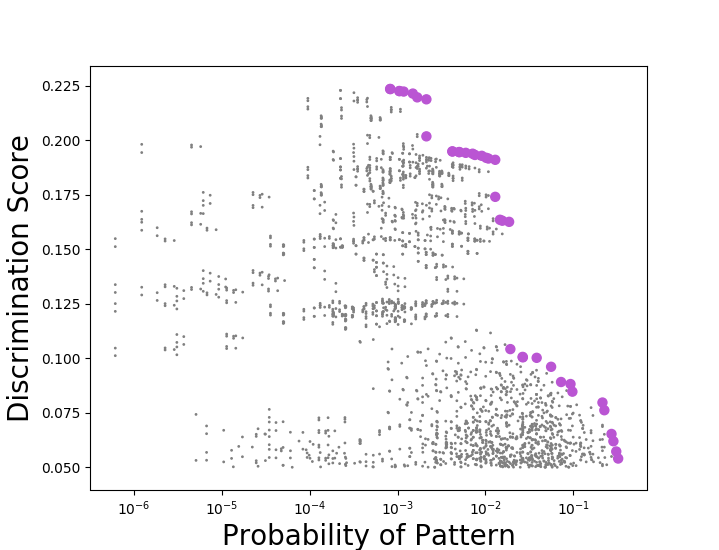}
\caption{Exact Pareto front}
\label{fig:pareto-exact}
\end{subfigure}
\begin{subfigure}{0.23\textwidth}
\centering
\includegraphics[width=0.98\columnwidth]{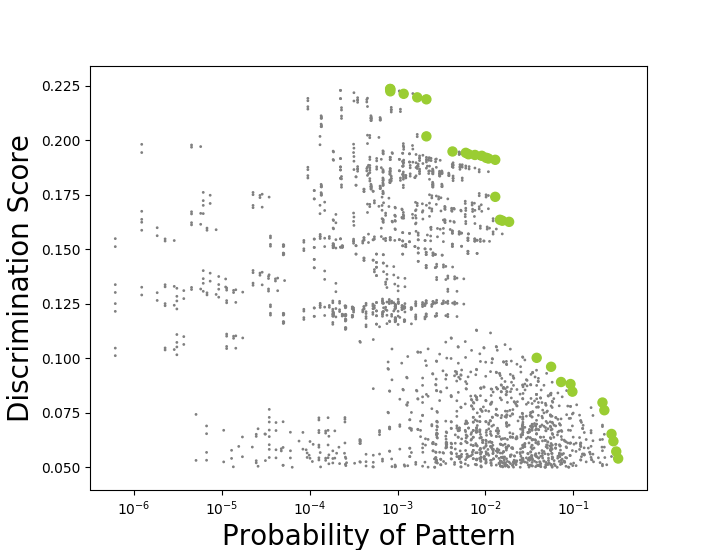}
\caption{Sampled Pareto front}
\label{fig:pareto-sample}
\end{subfigure}
\caption{Discrimination score and probability of all patterns (grey) and different summary patterns (color) for COMPAS.}
\label{fig:pareto_front}
\end{figure}

Lastly, we are interested in whether the patterns returned by the sampling algorithm are "interesting". To that end, we analyze the top 10 discrimination patterns produced by our sampling algorithm on COMPAS with a 3 second timeout and find that they indeed correspond to some of the most discriminating patterns in the model (\cref{fig:top10sampling}). Furthermore, we compare the Pareto optimal patterns from the sampling algorithm (\cref{fig:pareto-sample}) to the true Pareto front obtained through exact search (\cref{fig:pareto-exact}) and find that the sampling algorithm contains most of the same patterns in its Pareto front despite its short timeout. 
More concretely, the Pareto front from the sampling approach contains 30 patterns, of which 27 are in the true set of Pareto optimal patterns (there are 38 total).


\section{Conclusions, Limitations, and Discussion}
\label{sec:conclusion}
This paper studies fairness of probabilistic classifiers through the lens of discrimination patterns. With the goal of efficient and interpretable auditing for fairness, we introduce new classes of patterns, such as maximal, minimal, and Pareto optimal patterns, and propose efficient exact and approximate algorithms for mining these patterns in probabilistic circuits, an expressive class of tractable models. We empirically demonstrate the effectiveness of our methods in mining instances to help analyze a model's fairness properties.

Our method can be used by domain experts and ML practitioners not only for auditing in the deployment phase, but also as a subroutine when learning fair models. For example, one could use our approach to discover discrimination patterns, then enforce their elimination as a constraint during learning to obtain fair classifiers, which is a promising future direction.
Although we demonstrate the efficiency of our sampling-based mining algorithm empirically, we currently lack theoretical guarantees for the same, which we also leave as future work.
Lastly, we re-iterate that our method provides an efficient fairness auditing tool focusing on one specific notion out of many, and it remains the responsibility of domain experts and developers to interpret any pattern of unfairness in the context of the application.  

\section{Acknowledgments}
We thank the reviewers for their thoughtful feedback towards improving this paper. This work was funded in part by the DARPA Perceptually-enabled Task Guidance (PTG) Program under contract number HR00112220005, NSF grants \#IIS-1943641, \#IIS-1956441, \#CCF-1837129, Samsung, CISCO, a Sloan Fellowship, and a UCLA Samueli Fellowship.

\bibliography{main}

\begin{thebibliography}{39}
\providecommand{\natexlab}[1]{#1}

\bibitem[{Albarghouthi et~al.(2017)Albarghouthi, D'Antoni, Drews, and
  Nori}]{Albarghouthi17}
Albarghouthi, A.; D'Antoni, L.; Drews, S.; and Nori, A.~V. 2017.
\newblock FairSquare: Probabilistic Verification of Program Fairness.
\newblock 1(OOPSLA).

\bibitem[{Barocas and Selbst(2016)}]{barocas2016big}
Barocas, S.; and Selbst, A.~D. 2016.
\newblock Big data's disparate impact.
\newblock \emph{Calif. L. Rev.}, 104: 671.

\bibitem[{Bastani, Zhang, and Solar-Lezama(2019)}]{Bastani19}
Bastani, O.; Zhang, X.; and Solar-Lezama, A. 2019.
\newblock Probabilistic Verification of Fairness Properties via Concentration.
\newblock 3(OOPSLA).

\bibitem[{Bellamy et~al.(2018)Bellamy, Dey, Hind, Hoffman, Houde, Kannan,
  Lohia, Martino, Mehta, Mojsilovic, Nagar, Ramamurthy, Richards, Saha,
  Sattigeri, Singh, Varshney, and Zhang}]{Bellamy18}
Bellamy, R. K.~E.; Dey, K.; Hind, M.; Hoffman, S.~C.; Houde, S.; Kannan, K.;
  Lohia, P.; Martino, J.; Mehta, S.; Mojsilovic, A.; Nagar, S.; Ramamurthy,
  K.~N.; Richards, J.; Saha, D.; Sattigeri, P.; Singh, M.; Varshney, K.~R.; and
  Zhang, Y. 2018.
\newblock AI Fairness 360: An Extensible Toolkit for Detecting, Understanding,
  and Mitigating Unwanted Algorithmic Bias.

\bibitem[{Berk et~al.(2018)Berk, Heidari, Jabbari, Kearns, and
  Roth}]{berk2018fairness}
Berk, R.; Heidari, H.; Jabbari, S.; Kearns, M.; and Roth, A. 2018.
\newblock Fairness in criminal justice risk assessments: The state of the art.
\newblock \emph{Sociological Methods \& Research}, 0049124118782533.

\bibitem[{Choi and Darwiche(2017)}]{ChoiDarwiche17}
Choi, A.; and Darwiche, A. 2017.
\newblock On Relaxing Determinism in Arithmetic Circuits.
\newblock In \emph{Proceedings of the Thirty-Fourth International Conference on
  Machine Learning (ICML)}.

\bibitem[{Choi et~al.(2020)Choi, Farnadi, Babaki, and Van~den
  Broeck}]{choi2020learning}
Choi, Y.; Farnadi, G.; Babaki, B.; and Van~den Broeck, G. 2020.
\newblock Learning fair naive bayes classifiers by discovering and eliminating
  discrimination patterns.
\newblock In \emph{Proceedings of the AAAI Conference on Artificial
  Intelligence}, volume~34, 10077--10084.

\bibitem[{Choi, Vergari, and Van~den Broeck(2020)}]{ProbCirc20}
Choi, Y.; Vergari, A.; and Van~den Broeck, G. 2020.
\newblock Probabilistic Circuits: A Unifying Framework for Tractable
  Probabilistic Models.

\bibitem[{Chouldechova(2017)}]{chouldechova2017fair}
Chouldechova, A. 2017.
\newblock Fair prediction with disparate impact: A study of bias in recidivism
  prediction instruments.
\newblock \emph{Big data}, 5(2): 153--163.

\bibitem[{Chow and Liu(1968)}]{ChowLiu}
Chow, C.~K.; and Liu, C.~N. 1968.
\newblock {Approximating discrete probability distributions with dependence
  trees}.
\newblock \emph{IEEE Transactions on Information Theory}.

\bibitem[{Dang, Vergari, and Van~den Broeck(2022)}]{DangIJAR22}
Dang, M.; Vergari, A.; and Van~den Broeck, G. 2022.
\newblock Strudel: A Fast and Accurate Learner of Structured-Decomposable
  Probabilistic Circuits.
\newblock \emph{International Journal of Approximate Reasoning}, 140: 92--115.

\bibitem[{Darwiche(2003)}]{darwicheJACM-POLY}
Darwiche, A. 2003.
\newblock A Differential Approach to Inference in Bayesian Networks.
\newblock \emph{Journal of the ACM}, 50(3): 280--305.

\bibitem[{Darwiche and Marquis(2002)}]{darwiche2002knowledge}
Darwiche, A.; and Marquis, P. 2002.
\newblock A knowledge compilation map.
\newblock \emph{Journal of Artificial Intelligence Research}, 17: 229--264.

\bibitem[{Datta, Tschantz, and Datta(2015)}]{datta2015automated}
Datta, A.; Tschantz, M.~C.; and Datta, A. 2015.
\newblock Automated experiments on ad privacy settings: A tale of opacity,
  choice, and discrimination.
\newblock \emph{Proceedings on privacy enhancing technologies}, 2015(1):
  92--112.

\bibitem[{Dechter and Mateescu(2007)}]{DBLP:journals/ai/DechterM07}
Dechter, R.; and Mateescu, R. 2007.
\newblock {AND/OR} search spaces for graphical models.
\newblock \emph{Artif. Intell.}, 171(2-3): 73--106.

\bibitem[{Ding et~al.(2021)Ding, Hardt, Miller, and Schmidt}]{ding2021retiring}
Ding, F.; Hardt, M.; Miller, J.; and Schmidt, L. 2021.
\newblock Retiring adult: New datasets for fair machine learning.
\newblock \emph{Advances in Neural Information Processing Systems}, 34.

\bibitem[{Dua and Graff(2017)}]{Dua:2019}
Dua, D.; and Graff, C. 2017.
\newblock {UCI} Machine Learning Repository.

\bibitem[{Dwork et~al.(2012)Dwork, Hardt, Pitassi, Reingold, and
  Zemel}]{dwork2012fairness}
Dwork, C.; Hardt, M.; Pitassi, T.; Reingold, O.; and Zemel, R. 2012.
\newblock Fairness through awareness.
\newblock In \emph{Proceedings of the 3rd innovations in theoretical computer
  science conference}, 214--226. ACM.

\bibitem[{Feldman et~al.(2015)Feldman, Friedler, Moeller, Scheidegger, and
  Venkatasubramanian}]{Feldman15}
Feldman, M.; Friedler, S.~A.; Moeller, J.; Scheidegger, C.; and
  Venkatasubramanian, S. 2015.
\newblock Certifying and Removing Disparate Impact.
\newblock In \emph{Proceedings of the 21th ACM SIGKDD International Conference
  on Knowledge Discovery and Data Mining}, KDD '15, 259–268. New York, NY,
  USA: Association for Computing Machinery.
\newblock ISBN 9781450336642.

\bibitem[{Galhotra, Brun, and Meliou(2017)}]{Galhotra17}
Galhotra, S.; Brun, Y.; and Meliou, A. 2017.
\newblock Fairness Testing: Testing Software for Discrimination.
\newblock In \emph{Proceedings of the 2017 11th Joint Meeting on Foundations of
  Software Engineering}, ESEC/FSE 2017, 498–510. New York, NY, USA:
  Association for Computing Machinery.
\newblock ISBN 9781450351058.

\bibitem[{Ghosh, Basu, and Meel(2021)}]{Ghosh_Basu_Meel_2021}
Ghosh, B.; Basu, D.; and Meel, K.~S. 2021.
\newblock Justicia: A Stochastic SAT Approach to Formally Verify Fairness.
\newblock \emph{Proceedings of the AAAI Conference on Artificial Intelligence},
  35(9): 7554--7563.

\bibitem[{Ghosh, Basu, and Meel(2022)}]{Ghosh_Basu_Meel_2022}
Ghosh, B.; Basu, D.; and Meel, K.~S. 2022.
\newblock Algorithmic Fairness Verification with Graphical Models.
\newblock \emph{Proceedings of the AAAI Conference on Artificial Intelligence},
  36(9): 9539--9548.

\bibitem[{Griffiths et~al.(2010)Griffiths, Chater, Kemp, Perfors, and
  Tenenbaum}]{GRIFFITHS2010357}
Griffiths, T.~L.; Chater, N.; Kemp, C.; Perfors, A.; and Tenenbaum, J.~B. 2010.
\newblock Probabilistic models of cognition: exploring representations and
  inductive biases.
\newblock \emph{Trends in Cognitive Sciences}, 14(8): 357 -- 364.

\bibitem[{Hardt, Price, and Srebro(2016)}]{hardt2016equality}
Hardt, M.; Price, E.; and Srebro, N. 2016.
\newblock Equality of opportunity in supervised learning.
\newblock In \emph{Advances in neural information processing systems},
  3315--3323.

\bibitem[{Henderson et~al.(2015)Henderson, Herring, Horton, and
  Thomas}]{henderson2015credit}
Henderson, L.; Herring, C.; Horton, H.~D.; and Thomas, M. 2015.
\newblock Credit Where Credit is Due?: Race, Gender, and Discrimination in the
  Credit Scores of Business Startups.
\newblock \emph{The Review of Black Political Economy}, 42(4): 459--479.

\bibitem[{Holtzen, {Van den Broeck}, and Millstein(2020)}]{HoltzenOOPSLA20}
Holtzen, S.; {Van den Broeck}, G.; and Millstein, T. 2020.
\newblock Scaling Exact Inference for Discrete Probabilistic Programs.
\newblock \emph{Proc. ACM Program. Lang. (OOPSLA)}.

\bibitem[{Kisa et~al.(2014)Kisa, {Van den Broeck}, Choi, and
  Darwiche}]{KisaVCD14}
Kisa, D.; {Van den Broeck}, G.; Choi, A.; and Darwiche, A. 2014.
\newblock Probabilistic Sentential Decision Diagrams.
\newblock In \emph{Proceedings of the 14th International Conference on
  Principles of Knowledge Representation and Reasoning (KR)}.

\bibitem[{Koller and Friedman(2009)}]{koller2009probabilistic}
Koller, D.; and Friedman, N. 2009.
\newblock \emph{Probabilistic graphical models: principles and techniques}.
\newblock MIT press.

\bibitem[{Li et~al.(2021)Li, Zeng, Vergari, and Van~den Broeck}]{LiUAI21}
Li, W.; Zeng, Z.; Vergari, A.; and Van~den Broeck, G. 2021.
\newblock Tractable Computation of Expected Kernels.
\newblock In \emph{Proceedings of the 37th Conference on Uncertainty in
  Aritifical Intelligence (UAI)}.

\bibitem[{Liu, Mandt, and Van~den Broeck(2022)}]{LiuICLR22}
Liu, A.; Mandt, S.; and Van~den Broeck, G. 2022.
\newblock Lossless Compression with Probabilistic Circuits.
\newblock In \emph{International Conference on Learning Representations
  (ICLR)}.

\bibitem[{Madras et~al.(2018)Madras, Creager, Pitassi, and
  Zemel}]{madras2018fairness}
Madras, D.; Creager, E.; Pitassi, T.; and Zemel, R. 2018.
\newblock Fairness Through Causal Awareness: Learning Latent-Variable Models
  for Biased Data.
\newblock \emph{arXiv preprint arXiv:1809.02519}.

\bibitem[{Meila and Jordan(2000)}]{meila2000learning}
Meila, M.; and Jordan, M.~I. 2000.
\newblock Learning with mixtures of trees.
\newblock \emph{Journal of Machine Learning Research}, 1(Oct): 1--48.

\bibitem[{Nabi and Shpitser(2018)}]{nabi2018fair}
Nabi, R.; and Shpitser, I. 2018.
\newblock Fair inference on outcomes.
\newblock In \emph{Proceedings of the AAAI Conference on Artificial
  Intelligence}, volume~32.

\bibitem[{Peharz et~al.(2020)Peharz, Lang, Vergari, Stelzner, Molina, Trapp,
  Van~den Broeck, Kersting, and Ghahramani}]{PeharzICML20}
Peharz, R.; Lang, S.; Vergari, A.; Stelzner, K.; Molina, A.; Trapp, M.; Van~den
  Broeck, G.; Kersting, K.; and Ghahramani, Z. 2020.
\newblock Einsum Networks: Fast and Scalable Learning of Tractable
  Probabilistic Circuits.
\newblock In \emph{Proceedings of the 37th International Conference on Machine
  Learning (ICML)}.

\bibitem[{Poon and Domingos(2011)}]{poon2011sum}
Poon, H.; and Domingos, P. 2011.
\newblock Sum-product networks: A new deep architecture.
\newblock In \emph{2011 IEEE International Conference on Computer Vision
  Workshops (ICCV Workshops)}, 689--690. IEEE.

\bibitem[{Saleiro et~al.(2018)Saleiro, Kuester, Hinkson, London, Stevens,
  Anisfeld, Rodolfa, and Ghani}]{saleiro2018aequitas}
Saleiro, P.; Kuester, B.; Hinkson, L.; London, J.; Stevens, A.; Anisfeld, A.;
  Rodolfa, K.~T.; and Ghani, R. 2018.
\newblock Aequitas: A bias and fairness audit toolkit.
\newblock \emph{arXiv preprint arXiv:1811.05577}.

\bibitem[{Salimi et~al.(2019)Salimi, Rodriguez, Howe, and
  Suciu}]{salimi2019interventional}
Salimi, B.; Rodriguez, L.; Howe, B.; and Suciu, D. 2019.
\newblock Interventional fairness: Causal database repair for algorithmic
  fairness.
\newblock In \emph{Proceedings of the 2019 International Conference on
  Management of Data}, 793--810.

\bibitem[{Sonnenberg and Beck(1993)}]{sonnenberg1993markov}
Sonnenberg, F.~A.; and Beck, J.~R. 1993.
\newblock Markov models in medical decision making: a practical guide.
\newblock \emph{Medical decision making}, 13(4): 322--338.

\bibitem[{Vergari et~al.(2021)Vergari, Choi, Liu, Teso, and Van~den
  Broeck}]{VergariNeurIPS21}
Vergari, A.; Choi, Y.; Liu, A.; Teso, S.; and Van~den Broeck, G. 2021.
\newblock A Compositional Atlas of Tractable Circuit Operations for
  Probabilistic Inference.
\newblock In \emph{Advances in Neural Information Processing Systems 35
  (NeurIPS)}.

\end{thebibliography}

\clearpage

\appendix

\section{Proofs of Lemmas}

\rethm{lem:complete}{}{
    Let $P$ be a distribution over $D\cup\Z$ and $\x$ a joint assignment to $\X\subseteq\Z$. Also denote $\Vs = \Z\setminus\X$. Then for any $\Us \subseteq \Z\setminus\X$ the following holds:
    \begin{equation*}
        \max_{\us\in\val(\Us)} P(d \mid \x,\us) \leq \max_{\vs\in\val(\Vs)} P(d \mid \x,\vs)
    \end{equation*}
}
\begin{proof}
Consider any  $\Us \subset \Z\setminus\X$ and $W \notin \Us$. 
It suffices to show that
\begin{equation*}
        \forall\, \us\in\val(\Us),\quad P(d \mid \x,\us) \leq \max_{w\in\val(W)} P(d \mid \x,\us,w),
\end{equation*}
as the lemma then follows via a simple inductive argument. Denote $\val(W)=\{w_1,w_2,\ldots, w_n\}$. To show that there is at least one $w \in \val(W)$ such that $P(d\mid\x,\us) \leq P(d\mid\x,\us,w)$ for any $\us$, we will show that $P(d\mid\x,\us)>P(d\mid\x,\us,w_i)$ for $i=1,\dots,n-1$ implies that $P(d\mid\x,\us) < P(d\mid\x,\us,w_n)$.
First, for all $i \leq n-1$ we have:
\begin{align*}
    & P(d\mid\x,\us)>P(d\mid\x,\us,w_i) \quad \\
    &\implies  P(d,\x,\us) P(\x,\us,w_i)>P(\x,\us)  P(d,\x, \us,w_i).
\end{align*}
By taking the sum of both sides of the above inequality, we get:
\begin{align*}
    &\sum_{i=1}^{n-1} P(d,\x,\us) P(\x,\us,w_i) > \sum_{i=1}^{n-1} P(\x,\us)  P(d,\x,\us,w_i) \\
    &\implies  P(d,\x,\us) P(\x,\us)- \sum_{i=1}^{n-1} P(\x,\us)  P(d,\x,\us,w_i) \\ &\qquad > P(d,\x,\us) P(\x,\us)-  \sum_{i=1}^{n-1} P(d,\x,\us) P(\x,\us,w_i)\\
    &\implies  \frac{P(d,\x,\us)- \sum_{i=1}^{n-1} P(d,\x,\us,w_i)}{P(\x,\us)-   \sum_{i=1}^{n-1}  P(\x ,\us,w_i)} > \frac{P(d,\x, \us)}{P(\x, \us)}\\
    &\implies P(d\mid\x,\us,w_n)> P(d\mid\x,\us)
\end{align*}
\end{proof}

\rethm{lem:ratio}{}{
    Let $P$ be a distribution over $D\cup\Z$, $\x$ an assignment to $\X\subseteq\Z$, and $\Us\subseteq\Z\setminus\X$. Then,
    \begin{equation*}
        \argmax_{\us\in\val(\Us)} P(d \mid \x,\us) = \argmax_{\us\in\val(\Us)} \frac{P(\x,\us \mid d)}{P(\x,\us \mid \overline{d})}.
    \end{equation*}
}

\begin{proof}
     Since $P(d \mid \x, \us) = \frac{P(d,\x,\us)}{P(d,\x,\us)+P(\overline{d},\x,\us)} = \frac{1}{1+P(\overline{d},\x,\us) / P(d,\x,\us)}$, we obtain that
    \begin{align*}
         &\argmax_{\us\in\val(\Us)} P(d \mid \x,\us) =  \argmin_{\us\in\val(\Us)} \frac{P(\overline{d},\x,\us)}{P(d,\x,\us)} \\
         &=  \argmax_{\us\in\val(\Us)} \frac{P(d,\x,\us)}{P(\overline{d},\x,\us)}
         =  \argmax_{\us\in\val(\Us)} \frac{P(\x,\us \mid d)}{P(\x,\us \mid \overline{d})}.
    \end{align*}
\end{proof}

\section{Computing Upper Bounds}

\subsection{Discrimination Score}\label{sec:disc-bound}

Here we describe our algorithm to compute the upper bound on discrimination score (\cref{sec:search-algorithm}). Recall that we need to maximize or minimize quantities of the form $P(\x,\us\mid d) / P(\x,\us\mid \overline{d})$ over values of $\Us = \Z\setminus\X$ for some given evidence $\x\in\val(\X)$.
The pseudocode of our algorithm to maximize such ratio is given in \cref{alg:BRe}.
Again, we assume that the root of the PC is effectively a decision node on $D$, and thus its children represent the conditional distributions $P(\z\mid d)$ and $P(\z\mid\overline{d})$. Hence we can run \cref{alg:BRe} by giving those two children nodes as inputs.
Moreover, we can easily tweak the algorithm to minimize the ratio, by changing \cref{line:max} to return the minimum over non-zero values of the recursive calls if they exist, or zero otherwise.

The algorithm assumes PCs that satisfy two structural constraints: determinism and compatibility. 
A circuit is deterministic if the children of every sum node have disjoint supports (denoted by $\supp(n)$). In other words, for every complete assignment $\z$, at most one of the children nodes will have a non-zero output.
In addition, two circuits are compatible if they are: (1) smooth and decomposable; and (2) any pair of product nodes, one from each circuit, that are defined over the same set of variables decompose the variables in the same way. We refer the readers to~\citep{VergariNeurIPS21} for a more detailed discussion of compatibility.






\begin{algorithm}[tb]
    \caption{Best Ratio with Evidence: $\textsc{BR}(n,m)$
    }\label{alg:BRe}
    \begin{algorithmic}[1]
    \Input{deterministic and compatible PCs $n$ and $m$ over $\Z$; an assignment $\x\in\val(\X)$ for $\X\subset\Z$}
    \Output{$\max_{\us\in\val(\Us)} n(\x,\us)/m(\x,\us)$ where $\Us=\Z\setminus\X$}
    \If{$n,m$ are leaf nodes}
        \If{$\supp(n)\cap\supp(m)\not=\emptyset$ and $n(\x)\not=0,m(\x)\not=0$}
            \State{$\textsc{BR}(n,m)\gets 1$}
        \Else
            \State{$\textsc{BR}(n,m)\gets 0$}
        \EndIf
    \ElsIf{$n,m$ are product nodes} 
        \State{$\textsc{BR}(n,m)\gets \prod_{i=1}^{\abs{\ch(n)}} \label{line:prod} \textsc{BR}(n_i,m_i)$}
    \Else   \Comment{$n,m$ are sum nodes}
        \State{$\textsc{BR}(n,m)\gets \max_{n_i\in\ch(n),m_j\in\ch(m)} \frac{\theta_i}{\theta_j} \textsc{BR}(n_i,m_j)$} \label{line:max}
    \EndIf
    \end{algorithmic}
\end{algorithm}



\begin{proof}[Proof of Correctness]
We proceed via induction. For the leaves, as they are compatible, by definition their supports are either identical or completely disjoint. Thus, the maximum ratio is 1, 0, or undefined (we also propagate 0 in this case).

Next, consider two compatible product nodes. As they decompose the variables identically, we can order their children nodes such that $n(\z)=\prod_i n_i(\z_i)$ and $m(\z)=\prod_i m_i(\z_i)$, where $n_i$ and $m_i$ are over the same set of variables $\Z_i$. Let us write $\Us_i=\Us\cap\Z_i$ and $\X_i=\X\cap\Z_i$. Then, we have:
\begin{align*}
    \max_{\us\in\val(\Us)} \frac{n(\x,\us)}{m(\x,\us)}
    &= \max_{\us\in\val(\Us)} \frac{\prod_i n_i(\x_i,\us_i)}{\prod_i m_i(\x_i,\us_i)} \\
    &= \prod_i \max_{\us_i\in\val(\Us_i)} \frac{n_i(\x_i,\us_i)}{m_i(\x_i,\us_i)},
\end{align*}
leading to \cref{line:prod} in \cref{alg:BRe}.

Finally, consider two deterministic sum nodes. Then for any $\z$, at most one children each of $n$ and $m$ would evaluate non-zero values. That is, the sum nodes can effectively be treated as maximization nodes: e.g.\ $n(\z)=\sum_i n_i(\z)=\max_i n_i(\z)$. Moreover, among all pairs of children $n_i,m_j$, the ratio $n_i(\z)/m_j(\z)$ for any fixed $\z$ would be non-zero for at most one pair (again, we treat the ratio that is undefined as 0). Therefore, we have:
\begin{align*}
    \frac{n(\z)}{m(\z)}
    = \frac{\sum_i n_i(\z)}{\sum_j m_j(\z)}
    = \frac{\max_i n_i(\z)}{\max_j m_j(\z)}
    = \max_{i,j} \frac{n_i(\z)}{m_j(\z)}.
\end{align*}
Thus, we can break down the maximization as the following, corresponding to \cref{line:max}:
\begin{align*}
    \max_{\us\in\val(\Us)} \frac{n(\x,\us)}{m(\x,\us)}
    &= \max_{\us\in\val(\Us)} \max_{i,j} \frac{n_i(\x,\us)}{m_j(\x,\us)} \\
    &= \max_{i,j} \max_{\us\in\val(\Us)} \frac{n_i(\x,\us)}{m_j(\x,\us)}.
\end{align*}
%
\end{proof}



\subsection{Divergence Score}

\citet{choi2020learning} gives the following upper bound on the divergence scores of extensions of an assignment $(\x,\y)$:
\begin{align*}\label{eq:div-ub}
P(d,\x,\y) &\log \frac{\max _{\mathbf{z} \models \mathbf{x} \mathbf{y}} P(d \mid \mathbf{z})}{\min _{\mathbf{z} \models \mathbf{y}} P(d \mid \mathbf{z})} \\
&+ P(\overline{d} \mathbf{x} \mathbf{y}) \log \frac{\max _{\mathbf{z} \models \mathbf{x} \mathbf{y}} P(\overline{d} \mid \mathbf{z})}{\min _{\mathbf{z} \models \mathbf{y}} P(\overline{d} \mid \mathbf{z})},
\end{align*}
where $\z \models \x\y$ denotes a complete assignment to $\Z$ that agrees with $\x$ and $\y$ on their assignments to variables in $\X$ and $\Y$, respectively.

Observe that this upper bound once again requires efficient maximization and minimization of conditional probability of extensions. Thus, \cref{alg:BRe} allows us to leverage this upper bound and straightforwardly extend our exact search algorithm to mine divergence patterns as well.

\subsection{Relative Discrimination Score}
We define discrimination patterns using an absolute difference in conditional probabilities, but one may wish to characterize discrimination using a quantity that is proportional to the initial prediction probability. For instance, a prediction that goes from 0.15 to 0.05 after disclosing the sensitive attributes could be seen as more problematic than one that goes from 0.8 to 0.7, but they would have the same discrimination score.
We can alternatively define a score based on relative difference as the following.
\begin{defn}[Relative Degree of Discrimination]\label{def:relative-disc}
    Let $P$ be a probability distribution over $D \cup \Z$, and $\x$ and $\y$ be joint assignments to $\X\subseteq\Ss$ and $\Y\subseteq\Z\setminus\X$, respectively. The \emph{relative discrimination score} of pattern $\x,\y$, defined as $\Delta'(\x,\y)=\frac{ P(d \mid \x,\y)}{P(d \mid \y) }$. 
\end{defn}
We can mine discrimination patterns under this notion as well, by using a similar approach to derive the upper bound. That is, to get an upper bound on the relative discrimination score, we independently minimize/maximize $P(d \mid \x,\y)$ and $P(d \mid \y)$ over extensions, as described in \cref{sec:search-algorithm,sec:disc-bound}.

\section{Discussion of Summary Patterns}
The summary patterns are motivated theoretically as a means to capture the most interesting discrimination patterns to ML practitioners for performing quick and efficient audits. For instance, maximal and minimal patterns are intended to be a succinct way of explaining away exponentially many patterns or the lack thereof. Pareto optimal patterns attempt to account for probability of a pattern in addition to its discrimination score. 

To help the reader appreciate the quality of these patterns, in \cref{sec:summarizing} we provide concrete examples of maximal and minimal patterns for model on the COMPAS dataset. For instance, a single minimal pattern ($\delta = 0.1$) was able to represent and hence explain away 512 patterns (its extensions) out of 1164 total discrimination patterns in the model. In this case, the minimal pattern was $\x$ = \{Not Married\} and $\y$ = \{Low–Medium Supervision Level\}. This tells us that no matter what other features are observed, the corresponding group of people would experience unfair treatment from the model; that is, minimal patterns identify the root of discrimination by representing exponentially many patterns. This sort of succinct information is clearly helpful to an ML practitioner auditing the model for fairness.  As another example, recall  that if $\x\y$ is a maximal pattern, then no extension of it is a pattern. Hence, an individual with attributes $\x$ and $\y$ who may see a discrimination in the decision by disclosing their sensitive information would no longer receive such treatment if they additionally share other features. On a PC trained on COMPAS, we observe that among the 74 maximal patterns for $\delta=0.05$, none of them includes an assignment to the variable regarding ‘supervision level’, suggesting that there are many instances where an individual would not see an unfair prediction if the supervision level is additionally known.

Our proposed summary patterns are a first attempt to capture a small subset of patterns that would be most helpful to practitioners, and we leave it to future work to explore other possible types of summaries.

\section{Recovering Minimal Patterns}
Suppose we are given a set of patterns $\Sigma$. While a trivial algorithm to extract minimal patterns is quadratic in the number of patterns,  we can recover the minimal patterns in time $\mathcal{O}\left(P\cdot N^2+PNlog(PN)\right)$, where $P$ is the number of potential patterns and $N=|\Z|$. Note that this is particularly of interest when the number of patterns is very large.

First, we pre-process the patterns to identify candidate minimal patterns, which are patterns all of whose extensions are also patterns. Note that we get this for for free in the case of exact search with minor modifications. Consider the poset of all possible assignments $\x\y$ ordered by inclusion. We traverse this graph in level order, while maintaining a queue of nodes to visit and a set of assignments that we do not want to visit $S$. At the beginning of each level, we expand the nodes in $S$. Then, for each node in the queue for the current level that is not in $S$, if it is a candidate minimal pattern, we mark it as minimal, and add all its children to $S$. Otherwise, we add every child not in $S$ to the queue for the next level. 

\section{Sampling with Memoization}

At a high level, there are key two additions to \cref{alg:sampling} in \cref{alg:sampling-memo}.

First, the weights to sample from immediate extensions are no longer merely their discrimination scores. Instead, we maintain an estimator $\Phi(\x,\y)$ at each assignment corresponding to the expected discrimination score of an extension of $(\x,\y)$ (not just the immediate extensions by a single variable). $\Phi(\x,\y)$ is initialized as the discrimination score for every assignment $\x,\y$. After each sampling run (which refers to the complete extension path taken from the empty assignment $(\{\},\{\})$ to a complete assignment), we backtrack to update $\Phi(\x,\y)$ with our new information about average score of pattern encountered on the path after that assignment. More precisely, one can think of $\Phi(\x\y)$ as tracking the average discrimination score of a path of extensions from $(\x,\y)$, averaged over all extension paths explored so far. Observe that in contrast to \cref{alg:sampling}, consecutive sampling runs are no longer independent, and later runs have a more informed heuristics for exploring the search space.

Second, at any particular assignment $(\x,\y)$, the extensions are no longer directly sampled in proportion to $\Phi(\x,\y).$ Instead, we instead introduce a power factor of $\Gamma(\x,\y)=\left( 1+\frac{|\x|+|\y|}{n}\right)$ as a heuristic for how strongly we wish to adhere to our estimator in picking our path. One can view this as a control for exploration versus exploitation as $\Gamma(\x,\y)$ varies from 1 to 2. Intuitively, we are more open to exploration early on in our path as given a target pattern $(\x',\y')$, there are initially exponentially many paths to reach it. However, we prefer to exploit our estimator $\Phi(\x,\y)$ as the number of extension paths to $(\x',\y')$ reduces later in the sampling run. We note that this is particularly of significance in settings where the number of variables (and consequently the search space) is large, as for most practical purposes we are interested in quickly finding the most interesting patterns, and not necessarily interested in exploring the search space to extract all possible patterns.

\begin{algorithm*}[!t]
    \caption{$\textsc{Sample-Disc-Patterns}(\PC,\Z)$} \label{alg:sampling-memo}
    \begin{algorithmic}[1]
    \Input{a PC $\PC$ over variables $D\cup\Z$ and a threshold $\delta$}
    \Output{a set of sampled discrimination patterns $\Sigma$}
    \State{$\Sigma \leftarrow \{\}$}
    \State{$\Phi(\x,\y)\leftarrow\Delta(\x,\y) \quad \forall \x\y$}
    \State{$\sigma(\x,\y)\leftarrow 1 \quad \forall \x\y$}
    \Repeat \Comment{generate samples until timeout}
        \State{$(\x,\y)\leftarrow(\{\},\{\})$}
        \State{$ p \leftarrow [] $}
        \While{$\abs{\x}+\abs{\y} < n$}
            \For{$(\x',\y') \in \mathsf{extensions}(\x,\y))$}  \Comment{extensions by a single variable}
            \IfThen{$\Delta(\x',\y')>\delta$}{$\Sigma \leftarrow \Sigma \cup \{(\x',\y')\}$}
            \EndFor
            \State{$(\x,\y) \gets \mathsf{sample}_{\text{weight}:\Phi(\x,\y)^{\left( 1+\frac{|\x|+|\y|}{n}\right)}}(\mathsf{extensions}(\x,\y))$} 
            \State{$p \leftarrow p+(\x,\y)$}
        \EndWhile
        \For {$(\x,\y) \in \mathsf{reversed}(p))$} \Comment{update estimates}
        \State{$t \gets\frac{\Sigma_{i=|\x|+|\y|}^{n}\Phi(p[i])}{n-|\x|-|\y|}$}
        \State{$\sigma(\x,\y)\gets \sigma(\x,\y)+1$}
        \State{$\Phi(\x,\y)\gets \frac{\Phi(\x,\y) \cdot (\sigma(\x,\y)-1)+t}{\sigma(\x,\y)}$}
        \EndFor
    \Until{timeout}
    \State{\Return $\Sigma$}
    \end{algorithmic}
\end{algorithm*}

\section{Time Complexity}
\begin{itemize}
    \item  The time complexity of the upper bound routine (Algorithm 3) is $\mathcal{O}\left(|C|^{2}\right)$ where $|C|$ is the size of the circuit.
    \item Algorithm 1 is a branch and bound search, with the worst-case time complexity $\mathcal{O}$ (Search Space * Upper Bound Computation $)=\mathcal{O}\left(2^{n} *|C|^{2}\right)$, where $n$ is the number of attributes.
    \item Algorithm 4 (and its basic version Algorithm 2) is a sampling based approach which is run till timeout. Each sample is extremely efficient because for each attribute that we add in our construction, we only need to perform a few feed-forward evaluations of the circuit for computing marginals, which is linear time in the size of the circuit, resulting in a time complexity of $\mathcal{O}(n *|C|)$
\end{itemize}

\section{Additional Experimental Results}

First, we report the log likelihood and number of nodes in the PCs learnt in our experiments (\cref{tab:learnt_pcs}). In addition, in \cref{tab:extended_results}, we report an extended version of the results reported in \cref{tab:timeout}. In particular, we compare the number and scores of patterns found by exact search and \cref{alg:sampling-memo} with a fixed timeout in various settings (dataset, type of score, and threshold). We find that \cref{alg:sampling-memo} consistently outperforms exact search in both number of patterns found and score of highest pattern found across different settings.

\begin{table}[h]
\caption{Log likelihood and number of nodes in learnt PCs for various datasets}
\label{tab:learnt_pcs}
\centering
\scalebox{0.8}{
\begin{tabular}{l|r|r}
\toprule

Dataset & Log likelihood & Number of nodes \\ \midrule
Compas  & -192194.49   & 89      \\ 
Income  & -201072.99   & 119      \\ 
Adult   & -974185.15 & 291     \\  \bottomrule
\end{tabular}}
\end{table}

\begin{table*}[tb]
\centering
\caption{Number of patterns and highest score of pattern found by exact search and sampling.}
\label{tab:extended_results}
\begin{subfigure}{\textwidth}
\centering
\scalebox{0.95}{
\begin{tabular}{lrrr}
\toprule
Score          & Delta & Exact & Sampling\\ \hline
Discrimination & 0.01  & 584   & \textbf{980}         \\
              & 0.05  & 347   & \textbf{751}         \\
              & 0.1   & 210   & \textbf{347}         \\ \hline
Divergence     & 0.01  & 586   & \textbf{1186}        \\
              & 0.05  & 577   & \textbf{1644}        \\ \bottomrule
\end{tabular}}
\scalebox{0.95}{
\begin{tabular}{lrrr}
\toprule
Score          & Delta & Exact  & Sampling\\ \hline
Discrimination & 0.01  & \textbf{0.2236} & 0.2226      \\
              & 0.05  & \textbf{0.2236} & 0.2230      \\
              & 0.1   & \textbf{0.2236} & 0.2230      \\ \hline
Divergence     & 0.01  & 0.0015 & \textbf{0.0071}      \\
              & 0.05  & 0.0002 & \textbf{0.0009}      \\ \bottomrule
\end{tabular}}
\caption{COMPAS dataset with a 3 second timeout}
\end{subfigure}

\begin{subfigure}{\textwidth}
\centering
\scalebox{0.95}{
\begin{tabular}{lrrr}
\toprule
Score          & Delta & Exact & Sampling\\ \hline
Discrimination & 0.01  & 953   & \textbf{2236}        \\
              & 0.05  & 209   & \textbf{1090}        \\
              & 0.1   & 3     & \textbf{225}         \\ \hline
Divergence     & 0.01  & 840   & \textbf{2366}        \\
              & 0.05  & 818   & \textbf{2179}        \\ \bottomrule
\end{tabular}}
\scalebox{0.95}{
\begin{tabular}{lrrr}
\toprule
Score          & Delta & Exact  & Sampling\\ \hline
Discrimination & 0.01  & 0.1076 & \textbf{0.1658}      \\
              & 0.05  & 0.1076 & \textbf{0.1658}      \\
              & 0.1   & 0.1076 & \textbf{0.1658}      \\ \hline
Divergence     & 0.01  & 0.0046 & \textbf{0.0100}      \\
              & 0.05  & 0.0004 & \textbf{0.0023}      \\ \bottomrule
\end{tabular}}
\caption{Income dataset with a 5 second timesout}
\end{subfigure}

\begin{subfigure}{\textwidth}
\centering
\scalebox{0.95}{
\begin{tabular}{lrrr}
\toprule
Score          & Delta & Exact & Sampling \\ \hline
Discrimination & 0.01  & 42467 & \textbf{127855}      \\
              & 0.05  & 37167 & \textbf{113763}      \\
              & 0.1   & 30982 & \textbf{99578}       \\ \hline
Divergence     & 0.01  & 35792 & \textbf{133780}      \\
              & 0.05  & 35292 & \textbf{130232}      \\ \bottomrule
\end{tabular}}
\scalebox{0.95}{
\begin{tabular}{lrrr}
\toprule
Score          & Delta & Exact  & Sampling \\ \hline
Discrimination & 0.01  & 0.6725 & \textbf{0.6935}      \\
              & 0.05  & 0.6725 & \textbf{0.6871}      \\
              & 0.1   & 0.6725 & \textbf{0.6844}      \\ \hline
Divergence     & 0.01  & 0.0317 & \textbf{0.2125}      \\
              & 0.05  & 0.0162 & \textbf{0.1098}      \\ \bottomrule
\end{tabular}}
\caption{Adult dataset with a 600 second timeout}
\end{subfigure}
\end{table*}

\end{document}